\documentclass[runningheads]{llncs}

\usepackage{eccv}

\usepackage{eccvabbrv}

\usepackage{graphicx}
\usepackage{booktabs}
\usepackage[accsupp]{axessibility}  

\usepackage[pagebackref,breaklinks,colorlinks,citecolor=eccvblue]{hyperref}

\begin{document}

\title{XAI-guided Insulator Anomaly Detection for Imbalanced Datasets} 


\author{Maximilian Andreas Hoefler\inst{1} \and Karsten Mueller\inst{1} \and Wojciech Samek\inst{1,2,3}}
\institute{Fraunhofer Heinrich Hertz Institute, 10587 Berlin, Germany \and Technical University of Berlin, 10587 Berlin, Germany \and BIFOLD – Berlin Institute for the Foundations of Learning and Data, 10587 Berlin, Germany \\ \email{\{wojciech.samek,karsten.mueller\}@hhi.fraunhofer.de}}

\authorrunning{M.~Hoefler et al.}


\maketitle
\begin{abstract}
Power grids serve as a vital component in numerous industries, seamlessly delivering electrical energy to industrial processes and technologies, making their safe and reliable operation indispensable. However, powerlines can be hard to inspect due to difficult terrain or harsh climatic conditions. Therefore, unmanned aerial vehicles are increasingly deployed to inspect powerlines, resulting in a substantial stream of visual data which requires swift and accurate processing. Deep learning methods have become widely popular for this task, proving to be a valuable asset in fault detection. In particular, the detection of insulator defects is crucial for predicting powerline failures, since their malfunction can lead to transmission disruptions. It is therefore of great interest to continuously maintain and rigorously inspect insulator components. In this work we propose a novel pipeline to tackle this task. We utilize state-of-the-art object detection to detect and subsequently classify individual insulator anomalies. Our approach addresses dataset challenges such as imbalance and motion-blurred images through a fine-tuning methodology which allows us to alter the classification focus of the model by increasing the classification accuracy of anomalous insulators. In addition, we employ explainable-AI tools for precise localization and explanation of anomalies. This proposed method contributes to the field of anomaly detection, particularly vision-based industrial inspection and predictive maintenance. We significantly improve defect detection accuracy by up to 13\%, while also offering a detailed analysis of model mis-classifications and localization quality, showcasing the potential of our method on real-world data.
\end{abstract}
\section{Introduction}
The demand for green energy in its current form necessitates an increasing reliance on electrical power grids. Increased usage, e.g., of electric vehicles is estimated to increase the global electricity consumption by 11-20\%, with demand peaks additionally applying stress to the power grid \cite{energy_grid}. The stability of electrical grids and their continuing safe operation will thus be of paramount importance in the future. It follows that the health and functionality of power lines must be continuously monitored to ensure proper functioning and swift repairs in case of irregularities or damages. However, current inspection of power lines is arduous, labor intensive and slow. An automated, remote, and accurate inspection is therefore highly desirable by power suppliers and consumers alike. 

In recent years automated inspection and fault detection with the aid of deep learning methods has emerged as a promising solution \cite{Survey2023}. The aim is to, together with the assistance of unmanned aerial vehicles (UAVs), collect a large amount of visual data of powerline components and accurately determine damages or features which require attention and critical maintenance. Among the most important assets in  power lines is the insulator, which provides structural stability and simultaneously prevents the flow of current between conductors. Failures of insulators can be caused by a variety of factors such as precipitation, changing temperatures and wildlife interference which can results in transmission disruptions \cite{InsulatorBookRef}. Hence, sophisticated diagnostic methods which swiftly and reliably detect insulator anomalies are required.

In this work we propose a pipeline for insulator defect detection. Specifically, we generate bounding boxes for insulators and shells alike, thus significantly reducing the effect of misleading background features in images obtained from UAVs. We then train a YOLOv8 \cite{ultralytics} detection model, and subsequently separate the predicted insulators and shells from the original image. The shells are then categorized into three damage classes: healthy, flash-over pollution and broken, whereby the ratio of damaged to healthy samples is heavily skewed towards healthy shells. We train various architectures to classify the individual shells into the respective classes, however due to the imbalance of the dataset, the trained model is biased towards the classification of healthy components. To counteract the imbalance, the classification head is retrained using logistic regression on balanced sub-sampled partitions of the dataset. We then apply explainable AI methods, specifically Layer-wise Relevance Propagation (LRP) \cite{LRP}, to localize anomalies and provide explanations of model predictions. Finally, an analysis of the effect of image quality on the fault detection is conducted. To that end, we use a sharpness metric to discard images with significant motion blur in order to improve accuracy and guarantee a more robust anomaly detection.
\newline
Our contributions can be summarised as follows:
\begin{itemize}
    \item An insulator defect detection pipeline which not only detects entire insulators but also individual shells to minimize background influence
    \item An analysis of per class performance and worst class improvement using a re-weighting step to bias the model towards detection of defects
    \item A novel methodology of visualizing and localizing defects on insulator shells using LRP providing a pixel-wise localization of anomalies
    \item An analysis of model localization and an investigation of image mis-classification with focus on image sharpness and motion blur
\end{itemize}
Overall, our method specifically addresses class imbalance in insulator defect detection while using methods from explainable artificial intelligence (XAI) \cite{explainableAI} to locate damage types in insulator strings with state-of-the-art performance. Our method allows for a fine-grained analysis of defect detection, and provides insights into the effects of image quality and data imbalanced in the domain of predictive maintenance in industrial applications.
\label{sec:intro}
\section{Related Work}
\subsection{Insulator Defect Detection}
Interest and publication volume regarding deep learning and insulator defect detection has increased substantially over the past years \cite{Survey2023}. Although general machine learning and computer vision approaches have been used for defect detection in insulators \cite{CV4ID1, CV4ID2}, deep learning methods have proved to be a more flexible and accurate approach. Specifically, spurious attributes in the background of the image are the primary obstacles for accurate defect detection, along with large feature variability across different insulator types such as color, position and orientation \cite{Survey2023}. 
\newline
Defect detection methodologies can be broadly categorized into two different categories: parallel or multi-task and sequential methods \cite{Survey2023}. In multi-task methods, detection of insulators and damages occurs simultaneously by a single network. R-CNN, Faster-RCNN and YOLO architectures are among the most popular base methods, as used in \cite{cheng2019faster, YOLO, RCNN}. Multiple extensions exist, building on these base architectures. An example is MTI-YOLO as introduced in \cite{MTI-YOLO}, which uses spatial pyramid pooling (SPP) for aerial images accounting for the difficulty of detection in complex backgrounds. Another example can be found in \cite{RCNN-UNET} which combines the detection phase using a R-CNN network and a pixel classification using U-Net. 
\newline
In contrast, sequential defect detection first localizes the insulator, and subsequently detects the defective region. Generally, these methods perform slightly better due to reduced background interference, however there is also a larger computational effort involved in these tasks. \cite{TwoStageTao} introduces a two-stage defect detection cascade including a region proposal network (RPN) with a VGG16 backbone. This is followed by a defect detector network using a ResNet-101 inserting RPN and ROI Pooling layers between two sets of convolution layers. Alternatively, instead of an initial detection step, segmentation has also been used as in \cite{segment1, segment2}. This has the potential to improve accuracy, but is in return associated with an increased labelling effort. However, despite the vast literature there exist very few works which address the issue of class imbalance. Only a handful of methods such as \cite{PULEARN} address this issues, using positive unlabelled (PU) learning and focal loss to reduce the effect of class imbalance. This offers a promising approach to combat class imbalance, and can be extended using grid search (GS) and fixed threshold (FT) methods as introduced in \cite{PiGS} and \cite{PiFT}. We use these methods as a baseline comparison in the results.
\subsection{Datasets}
The aforementioned methods are applied to one of four publicly available insulator defect datasets. Most frequently encountered is the Chinese Powerline Insulator Dataset (CPLID) \cite{CPLID_data}, consisting of defective and non-defective insulators, whereby the defects are characterized by missing shells. Next, \cite{unifying} propose the Unifying Public Datasets for Insulator Detection (UPID) which adds to the existing CPLID dataset. Third, \cite{FINet_data} extends UPID with synthetic fogging augmentations dubbed synthetic foggy insulator dataset (SFID). These three datasets thus share a base set of images. In contrast, the Insulator Defect Image Dataset (IDID) \cite{IDID} is a separate dataset with three distinct classes namely: healthy, broken and flash-over pollution damages. This constitutes a greater challenge due to the fact that the damages only comprise a small portion of the overall image, and thus require a more fine-grained analysis. Moreover, few works have been published using this dataset. Among them is the multi-task approach as detailed in \cite{idid_paper} and a more broad approach for multiple datasets as outlined in \cite{idid_swissgrid}. In \cite{zhang2023idid} the authors also consider the IDID dataset, however they merge flash-over pollution damages and healthy insulators into a single non-defective category, only considering insulators with broken shells as defective. A very recent approach from \cite{HybridYOLO} utilizes a hybrid between YOLO and a Quasi-ProtoPNet to classify individual shells into the respective damage types. They additionally provide a localization mechanism of the defect site using a ProtoPNet architecture \cite{protopnet}, however their analysis focuses on overall model performance, and does not take individual class performance into account. 
\label{sec:related}

\section{Method}
\begin{figure*}[h]
  \centering
  \includegraphics[width=1\linewidth]{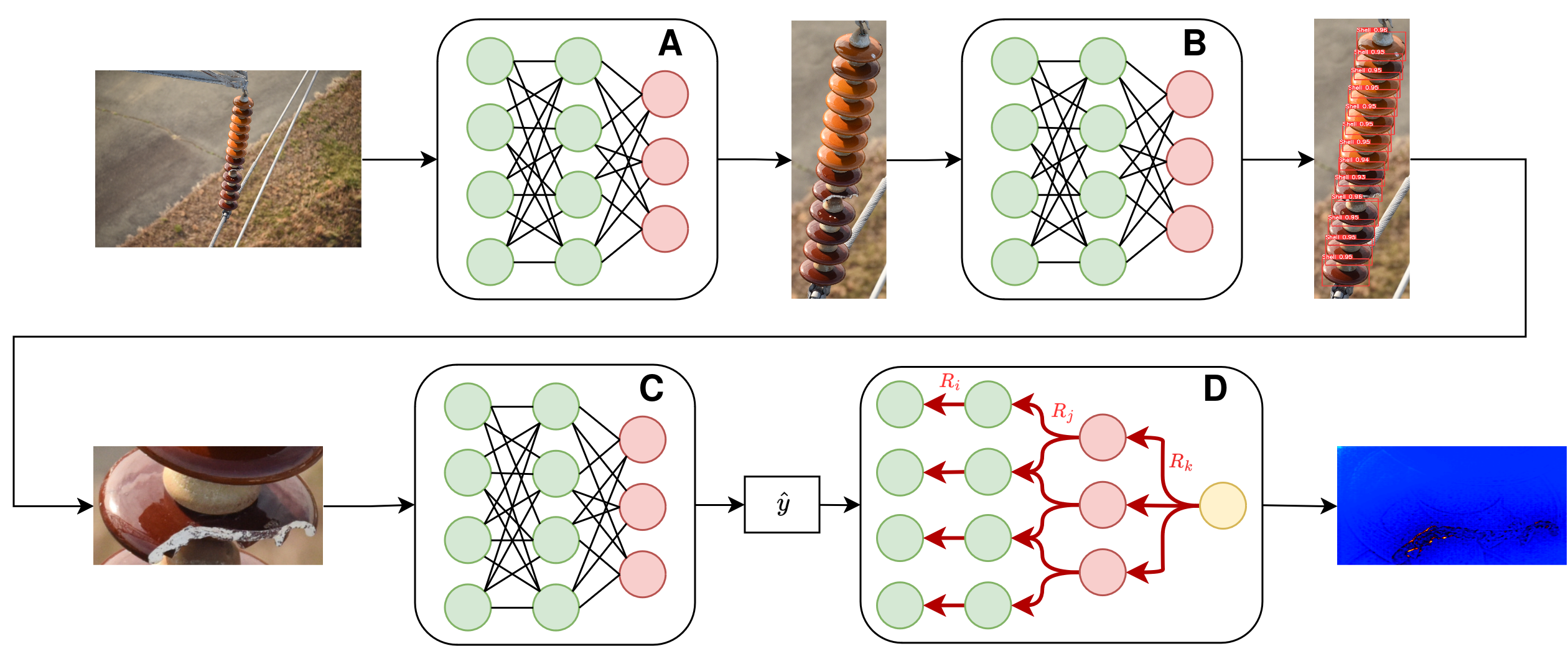}
  
  \caption{Process diagram showing our pipeline operating on an example insulator image. From left to right: \textbf{A} An input insulator image is processed by applying a YOLOv8 network $\mathcal{N}_I$ for whole insulator detection. Subsequently we perform a cropping operation $\mathcal{C}$ based on the predicted bounding box. \textbf{B} The individual shells are detected, via a YOLOv8 model $\mathcal{N}_s$ . The individual shells are cropped based on the predicted boxes via the cropping operator $\mathcal{C}$. \textbf{C} We perform classification using $\mathcal{N}_c$. \textbf{D} Heatmaps are generated for shells classified as "Damaged" using Layerwise Relevance Propagation, where the flow of relevance is indicated by the red arrows.}
  \label{fig:pipeline}
\end{figure*}
\subsection{Overview}
Our insulator defect detection pipeline \autoref{fig:pipeline} employs a two-stage object detection strategy to systematically identify and categorize insulator defects from UAV-captured images. The initial stage uses the dataset  $\mathcal{D} = \{(x_i, b_i) | (x_i, b_i) \in X \times B\}$, where $X$ comprises images of insulators and $B$ their corresponding bounding boxes. We utilize a neural network $\mathcal{N}_I$ to detect insulators within these images, generating bounding boxes $\hat{b}_i$. These boxes are then used to crop the original images, resulting in a set of insulator-focused images $X^c$.

In the second stage, we generate a specialized dataset $\mathcal{D}^s = \{(x^c_i, b_i^s) | (x^c_i, b^s_i) \in X^c \times B^s\}$, aimed at analyzing the condition of individual insulator shells. Here, $B^s$ represents the set of bounding boxes for each insulator shell. A corresponding network, $\mathcal{N}_s$, is applied to $\mathcal{D}^s$, producing bounding boxes $\hat{b}_i^s$ which are used to isolate individual shells from the original image. This yields a set of shell images $X^s$, which are subsequently classified into predefined defect categories (e.g., broken, flash-over polluted, healthy) based on the dataset $\mathcal{D}^r = \{(x_i^s, y_i) | (x^s_i, y_i) \in X^s \times Y\}$, where $Y$ denotes the set of defect condition labels. Finally layer-wise relevance propagation is applied to localize the damaged region on the individual shells by using the predicted label $\hat{y}$ of the classification network $\mathcal{N}_c$.
\subsection{Couteracting Imbalance via Finetuning}
 In the case of the IDID dataset, healthy shells are far more frequent than damaged shells, approximately in a ratio of $10:1$ of healthy to broken components and $5:1$ for healthy to flash-polluted shells. As a result, the classifier will be biased towards the majority class, favoring the features present in images of healthy components. However, especially in our application for reliable defect detection it is crucial to maximize the performance on defective components. In order to counteract the imbalance, we employ the method from \cite{freezing_layers} which retrains the final layer by running logistic regression on the model features obtained from the second-to-last layer. This method can be used to improve the worst-class performance, i.e., the class with lowest test accuracy. Specifically, we introduce individual class weights $w_c$ in the regression step, assigning a higher weight to faulty classes. 
To mitigate the imbalance further, we employ undersampling of the majority class, such that the ratio between class samples is approximately equal. This process involves dividing the dataset into ten partitions, $\mathcal{D}_p \subset \mathcal{D}$, each balanced through undersampling the majority class. For each partition, we extract a feature set $z^{l-1}_p = \mathcal{N}_c^{l-1}(x_i)$ from the second-to-last layer of the network, where $x_i$ belongs to a balanced partition $\mathcal{D}_p$. We thus obtain feature matrices of shape $|\mathcal{D}_p| \times F$, which are used as inputs to the logistic regression models, where $F$ is the number of features. For each partition, we derive a weight matrix $w_{i}$ as:
\begin{align}
    w_i = \underset{w}{\mathrm{argmin}} \, J(w, z^{l-1}_p, Y_p, w_c)
\end{align}
where $J$ is the cross-entropy loss objective and $Y_p$ the label set for the partition $\mathcal{D}_p$. The final model parameters $W$  are computed as the arithmetic mean value of the partition-specific weights. This refined logistic regression training strategy allows us to prioritize defect detection accuracy, specifically targeting the improvement of classification performance for the underrepresented classes.
\subsection{XAI-guided Anomaly Localization}
In order to obtain a localization of the damage type, we use Layerwise Relevance Propagation (LRP) \cite{LRP}. LRP follows a conservation principle, where the relevance received by a neuron must be distributed equally to the lower layer, whereby the general rule is applied:
\begin{align}
    R_j = \sum_k \frac{z_{jk}}{\sum_{j} z_{jk}} R_k
    \label{eq:lrp_eq}
\end{align}
Here, $z_{jk}$ represents how much neuron $j$ contributes to the relevance of neuron $k$, and $R$ the associated relevance. When applied to all neurons in the network, LRP maintains a layer-wise conservation property $\sum_j R_j = \sum_k R_k$, and globally, it satisfies the conservation property $\sum_i R_i = f(x)$, where $f(x)$ is the model's prediction. In the context of our work $f(x)$ specifies the neural network $\mathcal{N}_c$, where the predicted label $\hat{y}_i$ is used to generate a heatmap $\mathcal{H}(i,j)$. This heatmap represents the relevance $R$ of a pixel at position $(i,j)$ of an input image  $x_i$ in the testset $\mathcal{D}_T$.  It follows, that an accurate attribution, also necessitates an accurate prediction. If damage localization is a priority, then $\mathcal{N}_c$ should be able to discern damage types with high fidelity. This justifies the re-weighting scheme, as this improves classification accuracy on damaged components and hence improves the overall quality of heatmaps generated for faulty shells. 

The effectiveness of our explanation can be based on two general criteria, namely faithfulness and understandability. Faithfulness corresponds to an explanation which accurately represents the output neuron's behavior. In other words, it should faithfully reflect the model's decision process. This assumes that the neural network has correctly identified the relevant visual features for its predictions and has disregarded any irrelevant or adversarial factors in the input data. This assumption allows us to gauge how well the explanation aligns with the network's true reasoning process. In contrast, understandability relates to human interpretation, i.e., that explanations can reliably be understood and interpreted by human observers. For our application we strongly emphasize human understandability, since our model ultimately needs to justify its prediction and localization to users. Moreover, when inspecting a large amount of images it is important that the model's prediction of damages localize a precise area for maintenance to optimize the human intervention in repairs. In order to obtain an understandable explanation, one can choose from a variety of rules which dictate how the relevance $R_j$ of a neuron $j$ is computed. We refer to \cite{lrp_overview} for more details on the various rules that can be used. Additionally, we have the details and the used rules listed in the supplementary material. 
\subsection{Assessing Localization Quality}
As mentioned in the previous section, we want to produce understandable explanations of the model predictions. This requires an architecture and associated LRP-rule which maximizes human interpretability. To this end, we generate a set of segmentation masks for the damaged shells contained in the testset, in order to ascertain the quality of our localizations. We use the metric as introduced in \cite{TopKPaper}, where the authors define a pixel-wise intersection $tki$. In the context of our work, the metric is defined by a binary segmentation mask $M$ for the damaged region of a shell, and a binary mask $E$ of the top-$k$ features of heatmap $\mathcal{H}$. In other words the $k$ pixels of $\mathcal{H}$ which have the highest relevance $R$. The intersection is computed as follows:
\begin{align}
    tki= \frac{1}{k} \sum_{i=1}^{H} \sum_{j=1}^{W} (M_{i, j} \land E_{i, j})
    \label{eq:tki}
\end{align}
where $M$ and $E$ are the binary masks of size $W \times H$, $\land$ denotes the pixel-wise boolean AND operation and $k$ is number of pixels which are selected from $\mathcal{H}$. The metric attains a value of $tki=1$ if all top-$k$ pixels are contained within the segmentation mask region. Conversely, the metric is zero if no top-$k$ pixels are within the given region. We seek to find the combined network architecture and LRP rule, which maximizes the $tki$ score over the testset $\mathcal{D}_T$. This involves generating segmentation masks and $tki$ scores for each individual image in the testset and subsequently the average over all images. 
\subsection{Analyzing Mis-classification}
Finally, we conduct an analysis of the source of mis-classifcation. We attempt to determine why certain images are mis-classified or contain erroneous or poor localizations of damage type. To this end, we first analyze the relationship between sharpness and classification accuracy on the test data. We apply a sharpness analysis using a classical 2D high-pass or Laplacian operation as detailed in \cite{Laplacian_sharp} with the following kernel:
\begin{align}
  \text{K} = \frac{1}{6} \cdot \begin{bmatrix}
0 & -1 & 0 \\
-1 & 4 & -1 \\
0 & -1 & 0 \\
\end{bmatrix} 
\end{align}
Subsequently, we compute the Laplacian of our image, which approximates the second spatial derivative and enhances high spatial frequencies as strongly related to sharp edges. The luminance grey-scale input image $I$ from set $\mathcal{D}_T$ is then filtered:
\begin{align}
    L(x, y) = \sum_{i}^{\omega_x} \sum_{j}^{\omega_y} I(x+i, y+j) \cdot \text{K}(i, j)
\end{align}
where $\omega_x$ and $\omega_y$ correspond to the dimensions of a window centered at position $(i,j)$ in the image. Then, the variance of the filtered image is used as sharpness indicator:
\begin{align}
    V = \sum_{x=1}^{W} \sum_{y=1}^{H} [L(x, y) - \mu]^2
    \label{eq:sharpness_measure}
\end{align}
with
\begin{align}
    \mu = \frac{1}{W\cdot H}\sum_{x=1}^{W} \sum_{y=1}^{H} |L(x, y)|
\end{align}
With \autoref{eq:sharpness_measure} we are able to detect motion-blur, defocussing or blurriness in the images. These image properties are expected to have a negative impact on the classification accuracy, as features corresponding to damage types may not be sufficiently discerned. Therefore, we track the classification accuracy with respect to sharpness as defined in \autoref{eq:sharpness_measure} and define a sharpness threshold empirically based on the validation set performance. If an image falls below the threshold it is excluded from the inference step. 
\label{sec:method}

\section{Experiments and Results}
\subsection{Training and Architecture}
For insulator and shell detection we train medium-sized YOLOv8 networks using the ultralytics library \cite{ultralytics}. We train for a total of 300 epochs using an input image size of $512 \times 512$, and augment the dataset with rotations, horizontal and vertical flips, as well as shears, scales and translations. For all other experimental parameters, the default setting of the Ultralytics library for object detection are used. For classification, we tested MobileNetv2, ResNet18, ResNet50 and ResNet152 architectures pretrained on ImageNet. We used cross-entropy as the loss metric and SGD was used with a momentum value of 0.9, Nesterov acceleration and a learning rate scheduler. All details are listed in the supplementary material.
\subsection{Detecting Insulators}
\begin{figure}[t]
\centering
\includegraphics[width=0.7\linewidth]{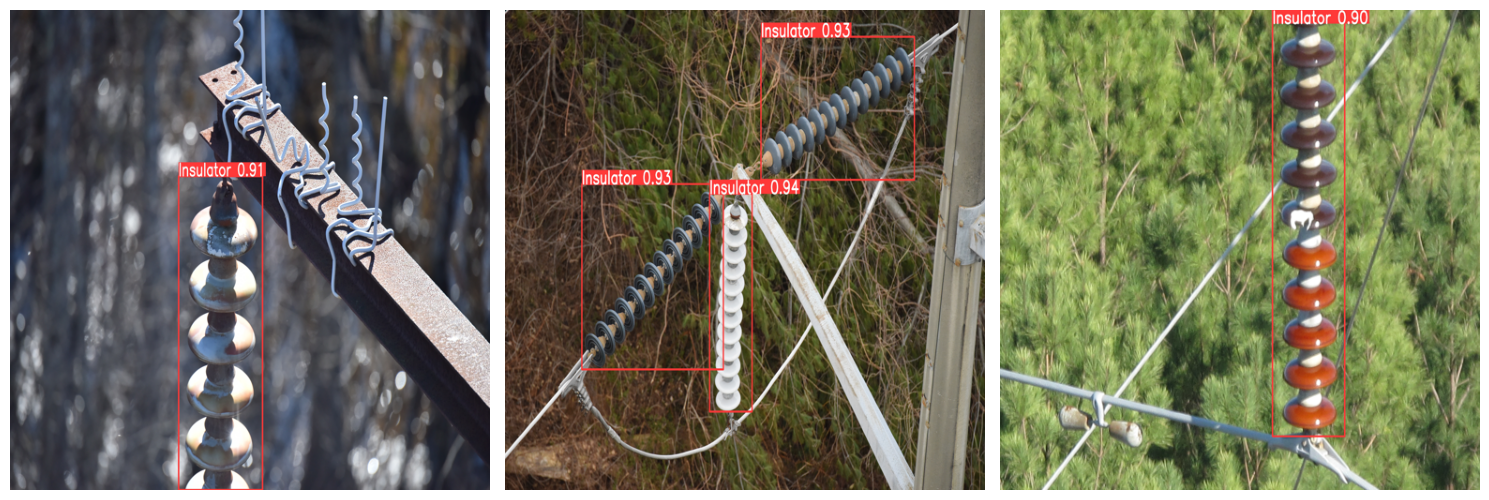}
\caption{Detection results on a sample of insulators from the test dataset using the training YOLOv8 model. One can see that insulators can be detected in a variety of backgrounds. Also multiple detection's are possible, including different colors, brightness and orientation. }
\label{fig:detection_samples}
\end{figure}

The results of the first object detection stage are shown in  \autoref{fig:detection_samples}. The figure shows that insulators are detected reliably. Overall, high accuracy results were obtained during training, achieving a mAP50 score of 0.97. This is in alignment with current state-of-the-art insulator object detection networks \cite{HybridYOLO}. 
In practice, the model is successful in separating all insulators in the source images, as required. Further evidence corroborating the success of training is shown in \autoref{fig:detection_samples}. Even insulators that are on the edges, obstructed or in complex backgrounds are easily detected with high fidelity.
\subsection{Detecting Shells}
\begin{figure}[h]
\centering
\includegraphics[width=0.7\linewidth]{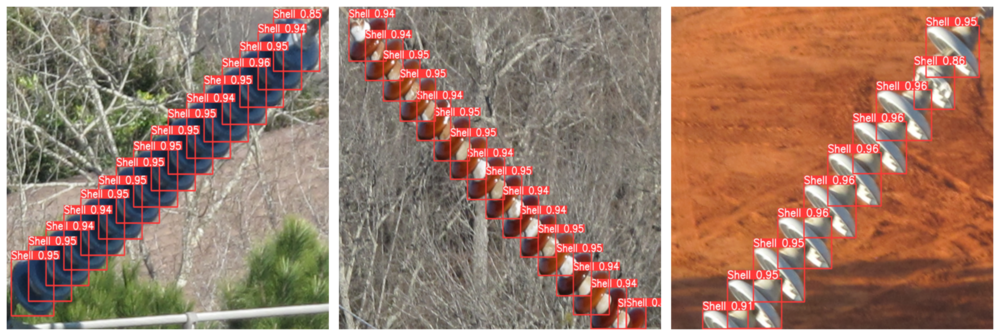}
\caption{Detection results on a sample of shells from the test dataset. One can see that shells are detected in a variety of environments, orientations and colors.
}
\label{fig:shell_detections}
\end{figure}

The separated insulator parts from the source image, are then used to identify individual shells. As mentioned above we use the same experimental setup for shell detection, thus providing a degree of continuity between these two stages. Again, we are able to separate individual shells with high accuracy, mAP50 of 0.98, which suggests that the task of detecting features corresponding to shells is readily achieved, even though the dataset contains a variety of colors, resolutions, lighting conditions and orientations. This is also competetive with current state-of-the-art such as \cite{HybridYOLO}. Furthermore, we can see the success of isolating shells by inspecting shell predictions of individual test images. \autoref{fig:shell_detections} shows a sample of images taken from the cropped test set. One can see that shells are identified accurately, even if these shells are damaged. This is crucial for our pipeline, as this allows for extracting healthy and broken shells alike. Therefore, the shell detection network is agnostic towards the detection of broken or healthy shells. 
\subsection{Classification}
The final step is the classification of shells. Tabulated in \autoref{tab:results} are the results for various standard deep classification models which were used to classify individual shells. Here, we focus on the per-class accuracy and compute the average in order to assess the quality of our model trained on an imbalanced dataset. As can be seen in the category of "Accuracy non-Reweighted", the accuracies are highly imbalanced, with worst-class performance being the Broken category across architectures with the exception of ResNet18. Since our goal is to ensure detection of defective classes, the initial results are not satisfactory, especially because the broken-damage type constitutes the greatest risk-factor for insulator failure and thus transmission breakdowns. 
\begin{table*}[t]\centering
\caption{Results for classification step. The column labelled as "Accuracy non-Reweighted" indicates the accuracy per class of the model trained on the shell dataset without accounting for the class imbalance. The second column labelled "Accuracy Reweighted" shows the results for the reweighting step. }
\begin{tabular}{c|cccc|cccc}
Model       & \multicolumn{4}{c|}{Accuracy non-Reweighted} & \multicolumn{4}{c}{Accuracy Reweighted}                 \\ \hline
            & Broken      & Flash     & Healthy     & Mean Acc.     & Broken         & Flash          & Healthy        & Mean Acc.      \\ \hline
MobileNetv2 & 0.836       & 0.900     & 0.952       & 0.896         & 0.918          & \textbf{0.921} & 0.882          & 0.904          \\
ResNet18    & 0.819       & 0.850     & 0.970       & 0.879         & 0.934          & 0.890          & 0.912          & 0.912          \\
ResNet50    & 0.836       & 0.870     & 0.976       & 0.894         & \textbf{0.934} & 0.920          & \textbf{0.937} & \textbf{0.930} \\
ResNet152   & 0.846       & 0.920     & 0.963       & 0.909         & 0.901          & 0.890          & 0.937          & 0.909         
\end{tabular}

\label{tab:results}
\end{table*}
\subsection{Re-biasing the Model}
\label{ssec:re_bias}
In order to bias the model towards higher accuracy on defective components, we leverage the fact that neural networks tend to learn general features in upper layers (closer to the input), and specific features towards deeper layers (closer to the output). Especially in our application this is true, due to the fact that the base object of interest, namely the insulator shell, is constant across the classes, with only the damage type or lack thereof being the variant. We adopt a retraining method inspired by \cite{freezing_layers}, where we perform logistic regression using the features generated up to the final linear layer of the network. In addition, we balance the dataset by removing a subset of the healthy-component class such that the final detection head will shift focus on defective components. 
The results are tabulated in \autoref{tab:results} in column "Accuracy Reweighted".  We primarily focus on increasing the classification accuracy with respect to the broken components, as these are most likely to cause powerline failures. We demonstrate that re-biasing is successful with the finetuning procedure. Although we sacrifice some of the accuracy of healthy components, we observe significant gains in both broken and flash damaged shells of up to 12\% and 7\%, respectively.
We find that ResNet50 offers the best overall performance. MobileNetv2 und ResNet18 demonstrate satisfactory performance for individual class accuracy, but are significantly less performant. Compared to the other architectures, the last linear layers of MobileNetv2 and ResNet18 are much smaller, only having dimensions $1240\times3$ and $512\times3$, in contrast to ResNet50 and ResNet152 which both contain a final layer of dimension $2048\times3$.  This could explain the performance gap, as more parameters are available for the logistic regression. 
Overall our results using the re-weighting technique are very promising and not only improve performance compared to standard training, but also allows for a degree of flexibility depending on the application. Specifically, the importance of specific classes can be elevated according to practical demands. 
As mentioned previously, very few works focus on per class accuracy or publish the results on per class accuracy scores. We compare our results to the limited studies which published accuracies or precision scores for individual classes in \autoref{tab:comparison}. One can see that our method outperforms current state-of-the-art on individual class accuracy. In particular, we make significant improvements in the categories Broken and Flash. Moreover, the gap between current work further corroborates that a per-class analysis is crucial in insulator defect detection. 
\begin{table}[t]
\centering
\caption{Comparison of Average Precision for each class across state-of-the-art methodologies. The first column refers to the precision of detecting whole insulator strings, and the columns Broken, Flash and Healthy refer to the damage type. The final column is the average of all values for the corresponding method.}
\begin{tabular}{ c|c|c|c|c|c}
    Method        & String         & Broken         & Flash          & Healthy        & Precision      \\ \hline
Impr.YOLOv7 \cite{YOLOv7IDID} & \textbf{98.60} & 82.62          & 79.61          & 94.20          & 88.76          \\
PU Learning \cite{PULEARN} & 93.14          & 88.06          & 81.03          & 90.20          & 88.11          \\
PU Learn FT \cite{PiFT}     & 92.37          & 80.72          & 90.56          & 88.08          & 87.93          \\
PU Learn GS  \cite{PiGS}   & 92.07          & 80.60          & 90.62          & 87.51          & 87.70          \\
YOLOv5 \cite{YOLO}     & 95.71          & 90.21          & 72.53          & 75.11          & 83.39          \\
CenterNet \cite{centernet}   & 86.35          & 71.42          & 53.44          & 52.10          & 65.83          \\
RetinaNet \cite{retinanet}  & 85.71          & 75.96          & 55.62          & 59.31          & 69.15          \\
Ours        & 97.21          & \textbf{93.08} & \textbf{92.45} & \textbf{91.40} & \textbf{93.54}
\end{tabular}
\label{tab:comparison}
\end{table}

\subsection{Localizing and Explaining}
\begin{figure*}[t]\centering

  \includegraphics[width=0.7\linewidth]{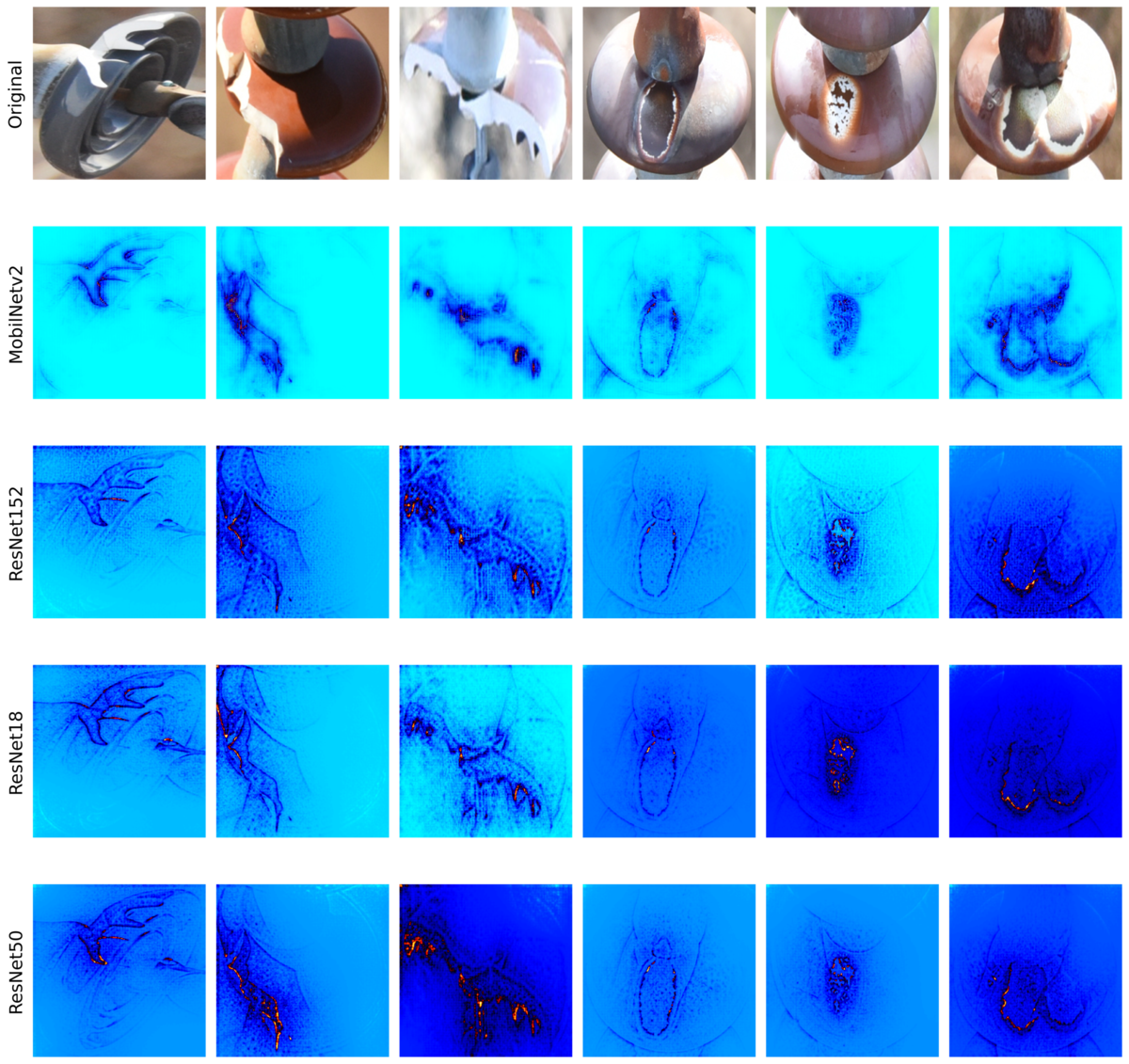}

  \caption{The results of our method showing sample heatmaps of shells with different damage types. The first row contains original images of damaged shells. Each row corresponds to a different architecture, comparing their performance. Darker pixels with a shift towards the red color spectrum indicate higher relevance.
    }
    \label{fig:XAI}
\end{figure*}

Although our classification model is competitive with current literature in detecting damages, the drawback is that, on its own, the classification model does not provide a localization mechanism. Object detection models have visualization baked into their architecture by design by utilizing bounding boxes to highlight a region of interest, however it is also possible to attain effective localization via XAI methods for classification models. We show that the damages can be more precisely localized in comparison to bounding box or other interpretability methods \cite{HybridYOLO}, as our approach yields a pixel-wise relevance. In \autoref{fig:XAI} we demonstrate this concept based on a sample of damaged shells.
We utilize LRP to propagate the predicted label and the associated relevance to the input image using the custom layer map as defined above. This results in a pixel-wise explanation where the model locates the damage in the form of a heatmap. \autoref{fig:XAI} shows the heatmaps of damaged insulator shells. Overall, LRP localizes defects very well, providing the highest relevance to the damaged region on the shell. Moreover, we observe a qualitative performance gap between individual models. Specifically, ResNet50 and MobileNetv2 perform best in terms of visual understandability, as they attribute the highest relevance to the correct region while simultaneously showing the least variance or scattering of relevance in the image. In contrast, ResNet152 and ResNet18 localize the correct region, but also highlight irrelevant regions to a larger degree. For example, both attribute significant relevance to the background surrounding the damaged shell of the first and second image. Moreover, in the third image the attribution is scattered around instead of showing a denser localization when compared to ResNet50 or MobileNetv2. 
Comparing ResNet50 and MobileNetv2, their localization performance is similar, however ResNet50 focuses more on relevant sites as indicated by the red and orange colors around the damaged areas. Based on visual inspection and classification performance, we conclude the ResNet50 performs best.
\subsection{Incorrect Classification and Explanation}
 
Although our pipeline yields accurate localizations, we also investigated the source of incorrect classifications and localizations. Specifically, we investigated heatmaps generated by out-of-focus or blurry images for classification performance. Here, the focus of the image is critical in obtaining accurate predictions as well as interpretable localizations. 
We additionally find that blurry images produce heatmaps with poor localizations. The attribution is scattered and not focused on relevant regions. This suggest that the model may only detect spurious attributes within the image which do no actually correspond to a localized damage type, even if the classification is correct. Heatmaps of blurry images can be found in the supplementary material. 

Based on these findings we recommend that images are checked for their degree of sharpness and accordingly mark or exclude images which do not pass a certain threshold, which we identified in this work using the described Laplacian filter. 
\begin{figure}[t]
\centering
  \includegraphics[width=0.6\linewidth]{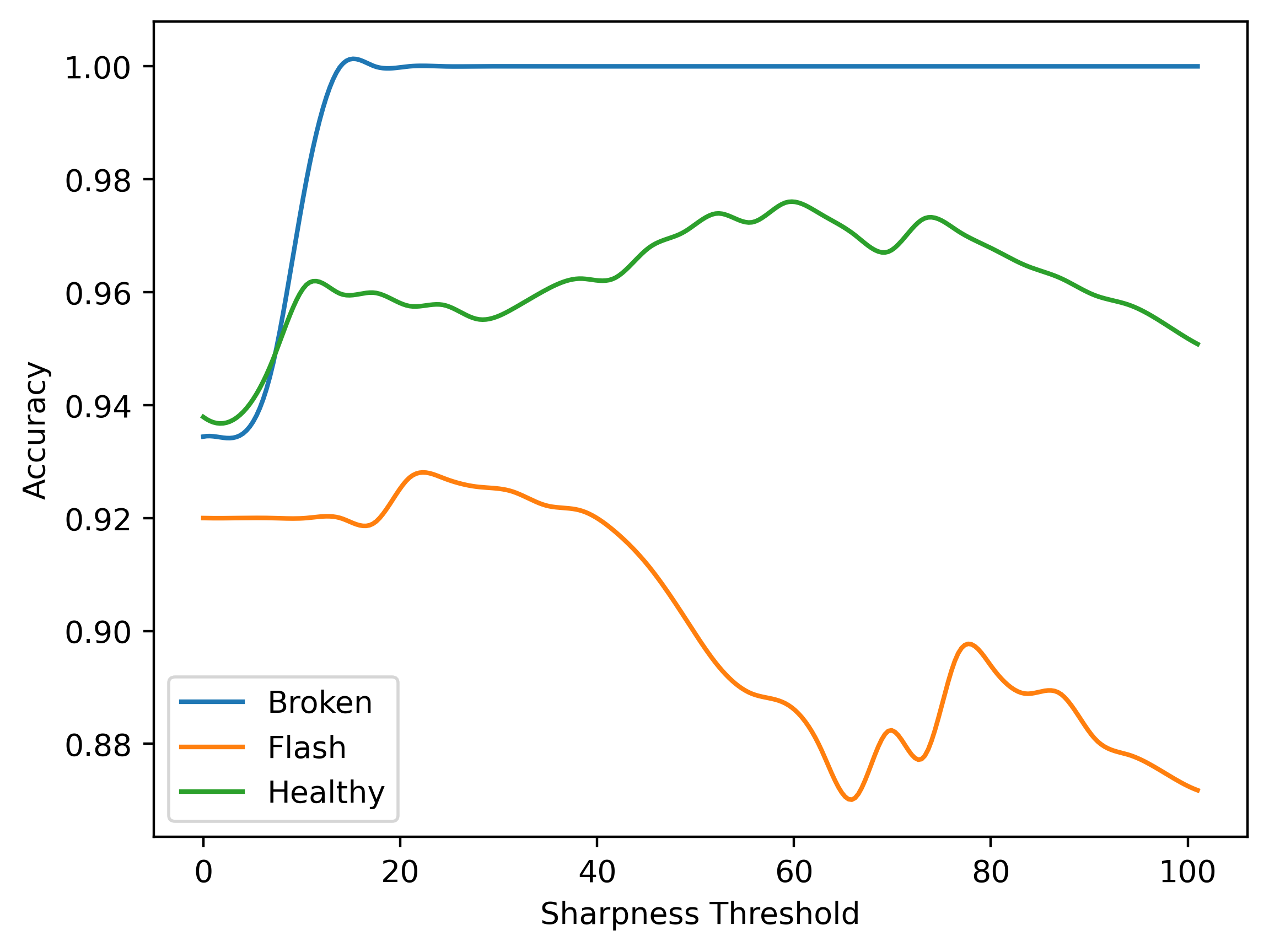}
  \caption{Sharpness versus prediction for the test dataset. We vary the threshold for the sharpness cutoff and plot the accuracy of each individual class accordingly.}
  \label{fig:sharp_vs_pred}
\end{figure}
\autoref{fig:sharp_vs_pred} shows the per class accuracy variation versus sharpness. It is evident that increasing the sharpness threshold in the beginning from 0 to 15, i.e., discarding the blurry and out-of-focus images below that threshold, improves the classification accuracy on broken components and at higher thresholds also flash-over polluted and healthy shells. Specifically, we obtain 100\% classification accuracy on broken components if the most blurry images are discarded. This model behavior is very desirable because broken components are the largest risk factor for insulator failures, and thus transmission disruptions. We find that with a sharpness threshold of 13.93 we obtain a macro-averaged accuracy of 96.8\%, with individual accuracies of 100\%, 94.84\% and 95.76\% for broken flash-over polluted and healthy, respectively. The threshold is determined using the validation set, but also by inspection of train images from dataset or an incoming test stream. The sharpness threshold can be adjusted according to the needs of the user and data availability.

\autoref{fig:sharp_vs_pred} also shows, that there is an inflection point for the healthy and flash-over pollution categories. We observe that the accuracy decreases above a certain sharpness threshold. This can be explained by the fact flash-over polluted images in the testset are generally in focus, and that increasing the threshold merely removes images which would otherwise be considered in focus. 
\subsection{Localization Quality}
 For assessing the quality of the generated heatmaps,  the top-$k$ metric defined in \autoref{eq:tki} is used as numerical measure. For this, different architectures and variations of the ResNet50 inference method are examined in \autoref{tab:tki_vs_architecture}. A score of 0 indicates that none of the top-$k$, i.e., $k$ most relevant features, are within the correct damage location, and a score of 1 indicates that all are within the target area. 
\begin{table}[tp]
\setlength{\tabcolsep}{3\tabcolsep}
\centering
\caption{Comparison of different variations of the ResNet50 architecture according to \autoref{eq:tki}. From top to bottom: the base architecture as trained on the imbalanced dataset, a re-weighted ResNet50 as detailed in \ref{ssec:re_bias}, a re-weighted ResNet50 with a sharpness threshold of 14.0 (S14) and one with a threshold of 30.0 (S30).}
\label{tab:tki_vs_architecture}
\begin{tabular}{c|c}
Variation            & $tki$-score    \\ \hline
Base        & 0.652          \\
Re-weighted     & 0.688          \\
S14-Re-Weighted & 0.670          \\
S30-Re-Weighted & \textbf{0.692}
\end{tabular}
\end{table}
The table shows that the re-weighting strategy significantly improves the localization quality, increasing the tki-score value from 0.652 to 0.688. We also observed that increasing the sharpness threshold at first decreases the top-$k$ score. This is explained by the fact that even an image which is out-of-focus, can posses a heatmap which localizes the damaged area correctly. The root cause is a mis-classification of the damage type. For example, when a broken component is classified as flash damaged or vice versa the correct area is highlighted despite an incorrect classification due to a feature similarity between flash-over polluted and broken damage types. Due to the similarity, classification accuracy and localization quality do not always correlate, and therefore heatmaps need to be additionally inspected against their class prediction. Above a particular sharpness threshold more flash damages appear to be classified as broken, thus resulting in a decrease in the per class accuracy in \autoref{fig:sharp_vs_pred}, but localization quality remains the same or even improves as the correct area is highlighted. 

\label{sec:results}

\section{Conclusion}
In this paper, we propose a novel insulator fault detection pipeline with pixel-wise localizations generated by LRP. In order to account for class imbalance we apply a re-weighting procedure to improve the classification performance on defective classes, paying special attention to broken components. For the first time, we address class imbalance in insulator detection from the vantage point of individual class accuracy, and apply XAI methods to render accurate damage localizations for insulator shell defects. Based on LRP, we also analyse failures of defect detection, revealing that blurry or generally unsharp images in the dataset degrade classification and effective localization. Moreover, we show that we can quantify the quality of images with respect to sharpness, and significantly improve the accuracy when applying a threshold to discard out-of-focus images. Lastly, we want to emphasize that our method can be transferred to wide range of applications, and is not limited to the domain of insulator defect detection. Therefore, future work will focus on additional domains requiring defect detection and localization. 
\label{sec:conclusion}
\newpage
\bibliographystyle{splncs04}
\bibliography{references}

%
%

\end{document}